%% file: main.tex
\documentclass[twoside,leqno,twocolumn]{article}

\usepackage[letterpaper]{geometry}

\usepackage{ltexpprt}
\usepackage{hyperref}
\usepackage[english]{babel}

\usepackage{amsmath}
\usepackage{amsfonts}
\usepackage{graphicx}
\usepackage{todonotes}
\usepackage{listings}
\usepackage{xcolor}
\usepackage{algorithm}
\usepackage[noend]{algpseudocode}
\usepackage{verbatim}
\usepackage{cleveref}
\usepackage{caption}
\usepackage{subcaption}

\definecolor{codegreen}{rgb}{0,0.6,0}
\definecolor{codegray}{rgb}{0.5,0.5,0.5}
\definecolor{codepurple}{rgb}{0.58,0,0.82}
\definecolor{backcolour}{rgb}{0.95,0.95,0.92}

\lstdefinestyle{mystyle}{
    backgroundcolor=\color{backcolour},
    commentstyle=\color{codegreen},
    keywordstyle=\color{magenta},
    numberstyle=\tiny\color{codegray},
    stringstyle=\color{codepurple},
    basicstyle=\ttfamily\footnotesize,
    breakatwhitespace=false,
    breaklines=true,
    captionpos=b,
    keepspaces=true,
    numbers=left,
    numbersep=5pt,
    showspaces=false,
    showstringspaces=false,
    showtabs=false,
    tabsize=2
}
\title{\Large Deep Learning and Spectral Embedding for Graph Partitioning}

\author{Alice Gatti \thanks{Computational Research Division, Lawrence Berkeley National Laboratory, Berkeley, CA} \and Zhixiong Hu\thanks{University of California, Santa Cruz, CA} \and Tess Smidt\thanks{MIT Department of Electrical Engineering and Computer Science, Cambridge, MA} \and Esmond G.~Ng\footnotemark[1] \and Pieter Ghysels\footnotemark[1]}
\date{}

\begin{document}
\maketitle


\begin{abstract} \small\baselineskip=9pt
We present a graph bisection and partitioning algorithm based on graph neural networks. For each node in the graph, the network outputs probabilities for each of the partitions. The graph neural network consists of two modules: an embedding phase and a partitioning phase. The embedding phase is trained first by minimizing a loss function inspired by spectral graph theory. The partitioning module is trained through a loss function that corresponds to the expected value of the normalized cut. Both parts of the neural network rely on SAGE convolutional layers and graph coarsening using heavy edge matching. The multilevel structure of the neural network is inspired by the multigrid algorithm. 
Our approach generalizes very well to bigger graphs and has partition quality comparable to METIS, Scotch and spectral partitioning, with shorter runtime compared to METIS and spectral partitioning.

\end{abstract}

\section{Introduction\label{sec:introduction}}

Graph partitioning is important in scientific computing, where it is used for instance to distribute unstructured data such as meshes (resulting from the discretization of partial differential equations) and sparse matrices over distributed memory compute nodes. When partitioning the graph corresponding to a sparse matrix, high quality partitions reduce communication volume and improve load balance in iterative solvers such as Krylov subspace iteration and multigrid. Graphs are also widely used in many social sciences where nodes and edges can represent a wide variety of subjects and their interactions. A problem closely related to partitioning is graph clustering. It is known that both graph partitioning and graph clustering are NP-hard, meaning that they are solved approximately with heuristic techniques, of which many have been developed. In general it is not known how far these approximations are from the optimal solution. Moreover, many of the heuristic algorithms are inherently sequential in nature, often exhibiting very irregular memory access patterns, and do not exploit current high-performance computing hardware efficiently since they rely mostly on integer manipulations.

We present an improvement to the deep learning framework for graph partitioning presented in~\cite{nazi2019gapICLR,nazi2019gap}, an unsupervised learning approach to graph partitioning referred to as GAP (generalized graph partitioning). The novel contribution of~\cite{nazi2019gapICLR} is the formulation of the loss function capturing the expected value of the normalized cut. However,~\cite{nazi2019gapICLR} focuses on applications where nodes in the graph have distinct features. Most of the graphs encountered in the context of scientific computing applications -- adjacency structures of sparse matrices, meshes from finite element/volume/difference discretizations, etc -- do not have node features. This is problematic for the approach from~\cite{nazi2019gapICLR}. We also observe that this approach, when used with graph convolutional networks with only message passing layers, does not generalize well. 

Our main innovation is in the structure of the neural network, which consists of two phases. The first phase, called embedding module, takes only the graph (i.e., the adjacency structure) as input, without any node or edge features. The output of this phase is a graph embedding, which is then passed to a second neural network, called the partitioning module. Two different loss functions are used to train the two separate parts of the network: first the embedding and then the partitioning module. The loss function for the embedding module is based on the graph Laplacian eigenvector residual, hence this module produces an approximate spectral graph embedding. This is necessary since our graphs do not have natural features. The second neural network uses the loss function from~\cite{nazi2019gapICLR}, which corresponds to the expected value of the normalized cut. The resulting partitioning algorithm consists of a single neural network, which combines both the embedding and partitioning modules, and outputs partition probabilities for each node for each partition. Both parts of the neural network use a multilevel structure that is inspired by the multigrid algorithm~\cite{briggs2000multigrid}.

The outline of the paper is as follows. \Cref{sec:problem} describes the problem and background. The algorithm is described in \cref{sec:algorithm}, with \cref{ssec:spectral_feature} and \cref{ssec:partitioning_module} introducing the embedding and partitioning modules, respectively. \Cref{sec:experiments} compares the new method to state-of-the-art graph partitioning codes and \cref{sec:conclusions} concludes the paper with thoughts on possible future work.

\section{Problem Definition and Background\label{sec:problem}}
Let $G = (V,E)$ be a graph, where $V = \{ v_i \}$ and $E = \{ e(v_i, v_j) \, | \, v_i \in V, \, v_j \in V\}$ are the sets of nodes and edges in the graph, respectively\footnote{The notation closely follows~\cite{nazi2019gapICLR}.}. We assume all graphs are undirected, meaning that if $e(v_i, v_j) \in E$, then $e(v_j, v_i) \in E$. Let $\mathcal{N}(v_i)$ denote the neighborhood of node $v_i$, i.e., all the nodes directly adjacent to $v_i$. The degree of a node is the size of its neighborhood. A graph $G$ can be partitioned into $g$ disjoint sets $S_1, S_2, \ldots, S_g$, where the union of the nodes in those sets is $V$ (i.e., $\cup_{k=1}^{g}{S_k = V}$) and each node belongs to one and only one set (i.e., $\cap_{k=1}^{g}{S_k = \emptyset}$), by simply removing edges connecting those sets.
The total number of edges that are removed from $G$ in order to form disjoint sets is called the edge cut, or simply the cut. For two partitions $S_k$ and $\bar{S}_k = V \setminus S_k$, the $\textnormal{cut}(S_k, \bar{S_k})$ is formally defined as
\begin{equation}
    \textnormal{cut}(S_k, \bar{S}_k) = \left | \{e(v_i,v_j)\ | \ v_i \in S_k,\ v_j \in \bar{S}_k \} \right | \, ,
\end{equation}
which can be generalized to multiple disjoint sets $S_1, S_2, \ldots, S_g$, with $\bar{S}_k$ the union of all sets except $S_k$:
\begin{equation}
    \textnormal{cut}(S_1, S_2, \ldots, S_g) = \frac{1}{2} \sum_{k=1}^{g}{\textnormal{cut}(S_k, \bar{S}_k)} \, . \label{eq:cut}
\end{equation}
One of the main objectives of graph partitioning is typically to minimize the cut. However, this needs to be done while balancing the partitions. A popular way to do this is to minimize the normalized cut instead:
\begin{equation}
    \textnormal{Ncut}(S_1, S_2, \ldots, S_g) = \sum_{k=1}^{g}{\frac{\textnormal{cut}(S_k, \bar{S}_k)}{\textnormal{vol}(S_k, V)}} \, , \label{eq:norm_cut}
\end{equation}
where the volume of a partition $S_k$
\begin{equation}
    \textnormal{vol}(S_k, V) = \left | \{e(v_i,v_j)\ |\ v_i \in S_k,\ v_j \in V \} \right | \, , \label{eq:volume}
\end{equation}
is the total degree of all nodes in $S_k$ in graph $G(V,E)$. This aims to minimize the cut while balancing the volumes of the partitions.

Other objectives are possible, such as minimization of the ratio cut \cite{chung1996spectral}
\begin{equation}
  \textnormal{Rcut}(S_1, S_2, \ldots, S_g) = \sum_{k=1}^{g}{\frac{\textnormal{cut}(S_k, \bar{S}_k)}{|S_k|}} \, , \label{eq:ratio_cut}
\end{equation}
instead of the normalized cut. The ratio cut tries to balance the partition cardinalities, instead of the volumes. Yet another possible objective is the conductance~\cite{bollobas2013modern}, also known as the quotient cut. The algorithms presented in \cref{sec:algorithm} can easily be adapted to these alternative objectives.

In sparse linear algebra applications, a partition of the adjacency structure of a sparse matrix corresponds to a set of rows and columns of the sparse matrix. The volume of the partition is the number of nonzero matrix elements in this set of rows/columns and the cut is the number of nonzero elements that have a row index in a different partition than its column index. When performing a sparse matrix-vector multiplication, the amount of work (floating point operations) per partition is proportional to the volume of the partition while the communication volume is proportional to the cut, which illustrates the importance of the normalized cut.

\subsection{Related Work\label{ssec:related}}
Popular and widely used graph partitioning codes are METIS~\cite{karypis1998fast} and Scotch~\cite{pellegrini1996scotch}. Both packages also have distributed memory implementations: ParMETIS~\cite{karypis1999parallel} and PTScotch~\cite{chevalier2008pt}, and METIS recently also gained a multithreaded variant called mtMETIS~\cite{lasalle2013multi}. Both METIS and Scotch use a multilevel graph partitioning framework, where the graph is first partitioned on a coarser representation (recursively) and the partitioning is then interpolated back and refined. Many heuristics can be used for the coarse level partitioning and the refinement. Popular choices are for instance Fiduccia-Mattheyses~\cite{fiduccia1982linear} and Kernighan-Lin~\cite{kernighan1970efficient}, diffusion~\cite{chevalier2006improvement}, Gibbs-Poole-Stockmeyer, greedy graph growing, etc.

The older package Chaco~\cite{hendrickson1993chaco} uses spectral graph theory for partitioning~\cite{pothen1990partitioning,simon1991partitioning}. This relies on the Fiedler vector~\cite{fiedler1973algebraic} $f$, which is an eigenvector of the graph Laplacian $L(G)$ (see \cref{ssec:spectral_feature} and \cref{ssec:spectralgraphpartitioning}) corresponding to the smallest non-zero eigenvalue. The Fiedler vector is typically very smooth, and can be used for partitioning by defining two partitions as $S_1 = \{ v_i \in V \, | \, f_i < c\}$ and $S_2 = \{ v_i \in V \, | \, f_i \geq c\}$, for some properly chosen value of $c$. The Cheeger bound~\cite{cheeger1969,chung1996spectral} guarantees that spectral bisection provides partitions with nearly optimal conductance (the ratio between the number of cut edges and the volume of the smallest part). Spectral methods make use of global information of the graph, while combinatorial algorithms like Kernighan-Lin and Fiduccia-Mattheyses rely on local node connectivity information.
Spectral methods have the advantage that they can be implemented efficiently on high performance computing hardware with graphics accelerators.

 The Mongoose graph partitioning code~\cite{davis2020algorithm} solves the discrete problem using a continuous quadratic programming formulation: $\textnormal{min}_{x \in \mathbb{R}^n}(1-x)^{\top} (A+I)x$, subject to $0 \leq x \leq 1$ and $\ell \leq 1^\top x \leq u$ with $\ell$ and $u$ bounds on the partition sizes.

Researchers have also tried to find inspiration in nature for the solution of NP problems. For instance \cite{Lucas_2014} uses an Ising model for magnetic dipole moments of atomic spins, which can be either up or down, $-1$ or $+1$, for the partitioning problem. Since atoms can sense the magnetic field of their neighbors, they will tend to line up with the same spin in order to minimize the overall energy of the system, much like neighboring vertices in a graph ``prefer'' to be in the same partition in order to minimize the graph cut.

\section{Algorithm Description\label{sec:algorithm}}
This section describes how a graph can be partitioned into $g$ disjoint sets $S_1, S_2, \ldots, S_g$ using a graph neural network.
The neural network will take as input a graph $G(V, E)$ and output a probability tensor $Y \in \mathbb{R}^{n \times g}$, where $Y_{ik}$ represents the probability that node $v_i \in V$ belongs to partition $S_k$. The probability that node $v_j$ does not belong to partition $S_k$ is $1 - Y_{jk}$. This network consists of two consecutive modules, an embedding module and a partitioning module. \Cref{ssec:spectral_feature} describes the embedding module and the corresponding loss function used to train it. \Cref{ssec:partitioning_module} introduces the partitioning module, with \cref{ssec:ncut_loss} discussing the loss function used to train this part of the network.

For the forward evaluation, the embedding is followed by a standardization operation and the partitioning module, and together are treated as a single neural network. 

The algorithm as presented here can be trained for partitioning the graph into an arbitrary, but fixed, number of partitions $g$. All experiments are for $g=2$, so-called graph bisection. For more general graph partitioning, one can recursively use graph bisection.

\subsection{Feature Engineering using Spectral Embedding\label{ssec:spectral_feature}}
Many of the graphs encountered in scientific computing applications do not have node or edge features, or often have edge features which are very similar between neighboring nodes. The neural network described in this section provides a technique, based on spectral graph theory, to compute a graph embedding to be used as node features. These features are then used by the partitioning module, \Cref{ssec:partitioning_module}, to partition the graph.

\begin{algorithm*}
	\textbf{Input:} graph $G(V, E)$ \\
	\textbf{Output:} feature tensor $F \in \mathbb{R}^{|V| \times 2}$
	\begin{algorithmic}[1]
		\Function{embedding}{$G$}
		    \State $\ell, \, G^0  \gets 0, \, G$
		    \While{$\textnormal{nodes}(G^{\ell}) > 2$} \Comment{$\textnormal{nodes}(G^{\ell})$ is number of nodes in $G$}
		        \State $G^{\ell+1}, \, I^{\ell+1} \gets \textnormal{coarsen}(G^{\ell})$ \hfill \Comment{$I^{\ell}$ holds interpolation info} \label{line:embedding_coarsening}
		        \State $\ell \gets \ell + 1$
            \EndWhile
		    \State $F^{\ell} \gets \begin{bmatrix} 1 & 0 \\ 0 & 1 \end{bmatrix}$ \hfill \Comment{initial features on coarsest graph} \label{line:embedding_Fcoarse}
		    \State $F^{\ell} \gets $ Tanh(SAGEConv$_{(\textnormal{coarse})}$($G^{\ell}, F^{\ell}$))
            \While{$\ell > 0$}
        		\State $F^{\ell-1} \gets \textnormal{interpolate}(F^{\ell}, G^{\ell}, I^{\ell})$
		        \hfill \Comment{interpolate $F^{\ell}$ using $G^{\ell}$ and $I^{\ell}$} \label{line:embedding_interpolation}
		        \State $\ell \gets \ell - 1$
		        \For{$t \gets 1 $ to $\nu$} \label{line:embedding_smoothing}
                    \State $F^{\ell} \gets $ Tanh(SAGEConv$_{(\textnormal{post}, t)}$($G^{\ell}, F^{\ell}$)) \hfill \Comment{smoothing of the features}
                \EndFor
            \EndWhile
            \For{$t \gets 1 $ to $\nu_{\textnormal{lin}}-1$}
                \State $F^0 \gets $ Tanh(Linear$_{(t)}$($F^0$)) \label{line:embedding_linear}
            \EndFor
            \State $F \gets $ Linear$_{(\nu_{\textnormal{lin}})}$($F^0$)\Comment{no activation of final Linear layer}
            \State \Return QR($F$) \Comment{orthogonalization of the output} \label{line:embedding_QR}
	    \EndFunction
	\end{algorithmic}
	\caption{Graph neural network implementation of the embedding module.}
	\label{alg:embedding_module}
\end{algorithm*}

\Cref{alg:embedding_module} shows the neural network for the embedding module, which is implemented using Pytorch Geometric~\cite{Fey_Lenssen_2019}. The only input is a graph $G(V, E)$, without node or edge features. The overall structure of this module is illustrated in \cref{fig:embedding_net}. The graph is coarsened repeatedly (\cref{line:embedding_coarsening}) until the coarsest representation only has $2$ nodes. The symbol $I^{\ell+1}$ denotes the mapping from nodes in $G^{\ell}$, the fine graph, to nodes in $G^{\ell+1}$, the coarse graph. This coarsening step uses the graclus~\cite{auer2012gpu} graph clustering code. Graclus implements a heavy edge matching to group neighboring nodes in clusters of size $2$ (or $1$). These clusters define the nodes for the coarse representation $G^{\ell+1}$ of $G^{\ell}$. A small number of nodes will remain unmatched, leading to clusters of size $1$. Hence, the coarsening rate is typically slightly less than $2$, resulting in $\ell^{\textnormal{max}}\sim \log{|V|}$ coarsening levels. Note that the first step in the graclus clustering algorithm is a random permutation of the nodes, which means that the graph coarsening phase is not deterministic.

At the coarsest level, where the graph $G^{\ell^{\textnormal{max}}}$ only has $2$ nodes, an initial feature tensor is constructed as the $2 \times 2$ identity matrix, \cref{line:embedding_Fcoarse}. This ensures that the features are linearly independent (orthogonal) and normalized. Then a single graph convolutional layer is applied to the combination of $G^{\ell^{\textnormal{max}}}$ and $F^{\ell^{\textnormal{max}}}$, followed by a nonlinear activation function.

The coarsest level feature tensor $F^{\ell^{\textnormal{max}}}$ is then repeatedly interpolated back to the finer graphs, using the clustering information recorded in $I^{\ell}$. After each interpolation step, a number $\nu_{\textnormal{post}}$ of convolutional layers is applied. There is a wide variety of graph convolutional layers described in the literature, and we have opted for the relatively simple SAGEConv layer~\cite{hamilton2017inductive}. The SAGE (SAmple and AGgregate) layer applied to a graph $G$ and an associated feature tensor $F \in \mathbb{R}^{|V| \times d}$ returns an updated feature tensor $F'$ with $F'_i \gets F_i W_1 + (\textnormal{mean}_{j \in \mathcal{N}(i)}F_j) W_2$, for $i \in \{1,\ldots,|V|\}$. Both weight matrices $W_1$ and $W_2$ belong to $ \mathbb{R}^{d \times d_{\textnormal{out}}}$, with $d_{\textnormal{out}}$ a tuning parameter. Each SAGE convolutional layer is followed by a hyperbolic tangent nonlinear activation function. We stress that the same convolutional layers (same weights) are used on each level. Hence, the number of trainable parameters in the network is independent of the size of the input graph.

After interpolation back to the original level $\ell = 0$, a number $\nu_{\textnormal{lin}}$ of linear layers are applied to the feature tensor $F^0$. The final step in the embedding module is a QR factorization, \cref{line:embedding_QR}, which ensures the features are orthogonal and have norm $1$.  The QR factorization is implemented using \texttt{torch.linalg.QR}, and gradients can back-propagate through this operation, which is differentiable in our context. 

\Cref{alg:embedding_module} is schematically illustrated in \cref{fig:embedding_net}.
\input{partition_fig}

\subsubsection{Eigenvector Residual as Loss Function\label{ssec::eigen_loss}}
We now discuss the loss function used to train the embedding module, \cref{alg:embedding_module}. The standard graph Laplacian is $L = (D - A)$, where $D$ is a diagonal matrix with the node degrees and $A$ is the binary adjacency matrix of the graph. The left normalized graph Laplacian $\tilde{L}$ -- also referred to as the random-walk normalized Laplacian -- is defined as
\begin{equation}\label{eq:left_norm_Laplacian}
\begin{split}
    \tilde{L}\left( G \right) =& D^{-1} \left( D - A \right) = I - D^{-1} A \\
    \tilde{L}_{ij} =& \begin{cases}
    \text{$1$} & \text{$i = j,\ \textnormal{degree}(v_i) \neq 0$,} \\
    -\frac{1}{\textnormal{degree}(v_i)} & \text{$i \neq j,\ v_j \in \mathcal{N}(v_i)$,} \\
    \text{$0$} & \text{otherwise}.
    \end{cases}
\end{split}
\end{equation}
Although $\tilde{L}(G)$ is not symmetric, it has real eigenvalues and it is positive semi-definite.
 Let its eigenvalues be ordered as $\lambda_1 = 0 \leq \lambda_2 \leq \dots \lambda_n$. The eigenvector corresponding to $\lambda_1 = 0$ is a constant vector, for instance the vector with all ones. If the graph is fully connected, then $\lambda_2 > 0$, while if $G$ is not fully connected, then the multiplicity of the eigenvalue zero corresponds to the number of fully connected components in the graph $G$.  The Fiedler vector~\cite{fiedler1973algebraic} $f$ is defined as the eigenvector of $\tilde{L}$ corresponding to the smallest non-zero eigenvalue, i.e., $\tilde{L}(G) f = \lambda_2 f$, assuming from here on that the graph is fully connected.

The Fiedler vector is of interest because it provides an approximate solution to the minimum normalized cut problem by solving the relaxed, continuous version of the problem \cite{shi2000normalized}. Hence, the Fiedler vector and other eigenvectors associated with small eigenvalues of $\tilde{L}$ are natural choices for feature vectors. However, computing these eigenvectors can be expensive. Hence, we train a neural network to find approximate eigenvectors in order to speed up the computation significantly. This is motivated by the fact that we do not need the exact eigenvectors, but we merely need a graph embedding to be used as input for a second neural network in \cref{ssec:partitioning_module}. 

We use \cref{alg:embedding_module} to compute $F = \begin{bmatrix} f_1 & \ldots & f_d \end{bmatrix} \in \mathbb{R}^{|V| \times d}$ by minimizing the loss function
\begin{equation}
    \mathcal{L}=\| \tilde{L}F - \Lambda F \|_2 + \sum_{i=1}^{d}{\lambda_i}  \label{eq:eig_loss}
\end{equation}
where $\Lambda = \textnormal{diag}(\lambda_1, \ldots, \lambda_d)$ and the approximate eigenvalues $\lambda_i$ are computed as the Rayleigh quotients $\lambda_i = F_{:i}^T \tilde{L} F_{:i}$. Note that the columns of the feature tensor $F$ are normalized. For all experiments, we set $d=2$, i.e., we compute the $2$ smallest eigenvalues (including $0$) and their corresponding eigenvectors, including the approximate Fiedler vector $f_2$. Since $f_1$ and $f_2$ are computed by a QR factorization, they are orthogonal and have norm $1$. Note that the exact $f_1$ is the constant vector with $(f_1)_i = 1/\sqrt{n}$, corresponding to $\lambda_1 = 0$, and the exact $f_2$ is orthogonal to $f_1$. Hence, the exact $f_2$ has zero mean: $\sum_{i=1}^{n}{(f_2)_i/\sqrt{|V|}} = 0$. However, the exact $f_2$ has standard deviation $\sigma = \sqrt{\sum{(f_2)_i^2}/|V|} = |V|^{-1/2}$. Hence, before passing the feature vector $f_2$ to the partitioning module, it is normalized as
\begin{equation}
    f_2 \gets \sqrt{|V|} \left( f_2 - \frac{\sum_i^{|V|}{(f_2)_i}}{|V|} \right) \label{eq:standardized_f2}
\end{equation}
such that it has $0$ mean and variance equal to $1$.

The approach presented here can be generalized to approximate multiple eigenvectors, for the smallest eigenvalues, instead of only the Fiedler vector. Passing more feature vectors to the partitioning module in \cref{ssec:partitioning_module} could in theory improve the graph bisection ($g=2$). Moreover, multiple eigenvectors are needed to successfully partition with $g > 2$~\cite{alpert1999spectral}.

\subsection{Graph Partitioning Module\label{ssec:partitioning_module}}

\begin{algorithm*}
	\textbf{Input:} graph $G(V, E)$, feature tensor $F \in \mathbb{R}^{|V| \times d}$ \\
	\textbf{Output:} partition probabilities $Y \in \mathbb{R}^{|V| \times g}$
	\begin{algorithmic}[1]
		\Function{partition}{$G$, $F$}
		    \State $\ell, \, G^0 \gets 0, \, G$
		    \State $F^0 \gets $ Tanh(SAGEConv$_{(\textnormal{first}, t)}$($G,F$))
		    \While{$\textnormal{nodes}(G^{\ell}) > 2$}
                \For{$t \gets 1 $ to $\nu_{\textnormal{pre}}$} \label{line:partition_pre_smoothing}
                    \State $F^{\ell} \gets $ Tanh(SAGEConv$_{(\textnormal{pre}, t)}$($G^{\ell}, F^{\ell}$)) \hfill \Comment{(pre) smoothing of the features}
                \EndFor
                \State $\tilde{F}^{\ell} \gets F^{\ell}$ \hfill \Comment{keep a copy of the features} \label{line:partitioning_saveF0}
                \State $G^{\ell+1}, \, I^{\ell+1} \gets \textnormal{coarsen}(G^{\ell})$ \hfill \Comment{$I^{\ell}$ holds interpolation info} \label{line:partition_coarsening}
		        \State $\ell \gets \ell + 1$
            \EndWhile
		    \For{$t \gets 1 $ to $\nu_{\textnormal{coarse}}$} \label{line:partition_coarse_smoothing}
                \State $F^{\ell} \gets $ Tanh(SAGEConv$_{(\textnormal{coarse}, t)}$($G^{\ell}, F^{\ell}$)) \hfill \Comment{smoothing of the features}
            \EndFor
            \While{$\ell > 0$}
        		\State $F^{\ell-1} \gets \textnormal{interpolate}(F^{\ell}, G^{\ell}, I^{\ell})$ \hfill \Comment{interpolate $F^{\ell}$ using $G^{\ell}$ and $I^{\ell}$} 
        		\State $\ell \gets \ell - 1$
		        \label{line:partition_interpolation}
		        \State $F^{\ell} \gets \left( F^{\ell}+ \tilde{F}^{\ell} \right) / 2$  \hfill \Comment{include original features} \label{line:part_addF0} \label{line:partitioning_averageFF0}
		        \For{$t \gets 1 $ to $\nu_{\textnormal{post}}$} \label{line:partition_post_smoothing}
                    \State $F^{\ell} \gets $ Tanh(SAGEConv$_{(\textnormal{post}, t)}$($G^{\ell}, F^{\ell}$)) \hfill \Comment{(post) smoothing of the features}
                \EndFor
            \EndWhile
            \For{$t \gets 1 $ to $\nu_{\textnormal{lin}} - 1$}
                \State $F^0 \gets $ Tanh(Linear$_{(t)}$($F^0$)) \label{line:partition_linear}
            \EndFor
            \State $F \gets $ Linear$_{(\nu_{\textnormal{lin}})}$($F^0$) \label{line:partition_linear_final} \hfill \Comment{no activation of final Linear layer}
            \State \Return softmax($F$, dim=1)
	    \EndFunction
	\end{algorithmic}
	\caption{Graph neural network implementation of the partitioning module.}
	\label{alg:partitioning_module}
\end{algorithm*}

The embedding vector, \cref{eq:standardized_f2}, computed in the previous section is passed to a second neural network, the partitioning module; see \cref{alg:partitioning_module} and the schematic illustration in \cref{fig:partition_net}. The partitioning module follows a similar multilevel flow as the embedding module, using the graclus clustering for graph coarsening. However, now there are three sets of convolutional layers: those applied before coarsening the graph and the associated features, those applied on the coarsest graph, and those applied after the interpolation. We would like to point out the similarity between the structure of this network and the multigrid algorithm. As in multigrid, we refer to the convolution before the coarsening as pre-smoothing, and to the convolution after the interpolation as post-smoothing. However, in contrast to multigrid, each convolutional layer is followed by a nonlinear activation function. For the pre and post-smoothing, $\nu_{\textnormal{pre}}$ and $\nu_{\textnormal{post}}$ layers are used, respectively. As for the embedding module, the same convolutional layers are used on each level.

At each level, the feature tensor is saved, \cref{line:partitioning_saveF0}, after applying the pre-smoothing and coarsening/interpolation. Then after the interpolation, this $\tilde{F}^{\ell}$ tensor is added to the interpolated tensor $F^{\ell}$, \cref{line:partitioning_averageFF0}. This is done because the coarsening can destroy much of the high frequency information, which cannot be represented on the coarser graph, available in the original tensor. Note the analogy with multigrid, where the coarse problem solves for the error using the coarsened residual and then subtracts the interpolated error from the original approximation.

After the final interpolation step, a number of dense layers are applied. Finally, a softmax layer outputs the $Y$ tensor with probabilities for each node and each partition. This is defined as a layer $\sigma$ that is applied for each node, such that the sum of the probabilities over the partitions is $1$: $\sigma(F_{:j}) = \exp(F_{:j}) / \sum_{j=1}^g \exp(F_{:j})$, where $g$ is the number of partitions. \Cref{alg:partitioning_module} is schematically illustrated in \cref{fig:partition_net}. We also implement \cref{alg:partitioning_module} using PyTorch Geometric~\cite{Fey_Lenssen_2019}.

\subsubsection{Expected Normalized Cut Loss Function\label{ssec:ncut_loss}}
Recall from \cref{eq:cut} that $\textnormal{cut}(S_k, \bar{S}_k)$ is the number of edges $e(v_i,v_j)$ with $v_i \in S_k$ and $v_j \notin S_k$. Since $Y_{ik}$  is the probability that $v_i \in S_k$ and $1 - Y_{jk}$ is the probability that $v_j \notin S_k$, we have
\begin{equation}
\begin{split}
    \mathbb{E}\left[ \textnormal{cut}(S_k, \bar{S}_k)\right]
     =& \sum_{i=1}^{|V|}{\sum_{v_j \in \mathcal{N}(v_i)}{Y_{ik} \left(1 - Y_{jk} \right)}}\\
     =& \sum_{i=1}^{|V|}{\sum_{j=1}^{|V|}{Y_{ik}(1 - Y_{kj}^T) A_{ij}}} \, .
\end{split}
\end{equation}
Let $D$ be a column vector with $D_i = \textnormal{degree}(v_i)$. Then
\begin{equation}
    \mathbb{E}\left[\textnormal{vol}(S_k, V) \right] = \Gamma_k \quad \textnormal{with} \quad \Gamma = Y^T D
\end{equation}
and hence
\begin{equation}
    \mathbb{E}\left[ \textnormal{Ncut}(S_1, S_2, \ldots, S_g) \right]
     = \sum{(Y \oslash \Gamma)(1 - Y)^T \odot A} \label{eq:exp_Ncut}
\end{equation}
where $\oslash$ and $\odot$ denote element-wise division and multiplication, respectively. The result of $(Y \oslash \Gamma)(1 - Y)^T \odot A$ is a $|V| \times |V|$ sparse matrix and the summation in~\Cref{eq:exp_Ncut} is over all its elements. Minimizing the loss function
\begin{equation}
    \mathcal{L} = \sum_{k=1}^{g}\sum_{i=1}^{|V|}{\sum_{j=1}^{|V|}{\frac{Y_{ik}(1 - Y_{kj}^T) A_{ij}}{\Gamma_{k}}}} \label{eq:gap_loss}
\end{equation}
is equivalent to minimizing the expected value of the normalized cut. \Cref{fig:gap_loss} in \cref{app:additional_material} shows how this can be efficiently implemented using PyTorch Geometric~\cite{Fey_Lenssen_2019}.

\Cref{eq:exp_Ncut} and the corresponding loss function \cref{eq:gap_loss} for the normalized cut appear in~\cite{nazi2019gapICLR}, which is published as a proceedings paper for ICLR 2019. All our experiments use \cref{eq:gap_loss} as the loss function. However, a similar article is available from the arXiv~\cite{nazi2019gap}, which uses this loss function but with an additional term to balance the cardinalities of the partitions:
\begin{equation}
    \mathcal{L} = \sum_{k=1}^{g}{\sum_{i,j=1}^{|V|}{\frac{Y_{ik}(1 - Y_{kj}^T) A_{ij}}{\Gamma_{ik}}}} + \sum_{k=1}^{g}\left( \sum_{i=1}^{|V|}{Y_{i,k}} - \frac{|V|}{g} \right)^2 \, , \label{eq:gap_loss_balanced}
\end{equation}
by forcing the cardinality of each partition to be close to $|V| / g$. Note that $\mathbb{E}[|S_k|] = \sum_{i=1}^{|V|}{Y_{ik}}$. All our experiments use \cref{eq:gap_loss} as the loss function.

\subsection{Spectral Graph Partitioning}\label{ssec:spectralgraphpartitioning}
As discussed in \cref{ssec:related} and \cref{ssec::eigen_loss}, the Fiedler vector $f$ provides a way to partition a graph such that the partitions are balanced and the cut is small~\cite{shi2000normalized}. Trivially, one can assign all vertices with a negative coordinate of the Fiedler vector to one partition, and those with a positive coordinate to the other. More generally, one can partition as $S_k = \{ v_i \in V \, | \, f_i < c\}$ and $\bar{S}_k = \{ v_i \in V \, | \, f_i \geq c\}$, for some properly chosen threshold value $c$, see also~\cite{pothen1990partitioning}. In order to get the best partitioning we compute the normalized cut for each $c \in \{f_{i}\ \vert\ 1 \leq i \leq |V|\}$ and pick the one that minimizes the normalized cut. In the following we will refer to this method as spectral graph partitioning when it is used with the exact Fiedler vector, or as approximate spectral partitioning when used with the approximate Fiedler vector as computed with the embedding module from \cref{ssec:spectral_feature}.

\section{Experiments\label{sec:experiments}}

All experiments were run on a desktop computer with an AMD Ryzen 9 3950X 16-core processor with 128GB of memory and a GeForce RTX 2060 SUPER GPU with 8GB GDDR6 memory.

For the embedding module, \cref{alg:embedding_module}, we set $\nu=2$, $\nu_{\textnormal{lin}}=4$, and used $32$ input and output channels for all $\textnormal{SAGEConv}_{(\textnormal{post},\cdot)}$ convolutional layers, $2$ input and $32$ output channels for the $\textnormal{SAGEConv}_{(\textnormal{coarse})}$ convolutional layer, $32$ input and $16$ output channels for Linear$_{1}$, $16$ input and $32$ output channels for Linear$_{2}$, $32$ input and output channels for Linear$_{3}$ and $32$ input and $2$ output channels for Linear$_{4}$. The total number of trainable parameters in the embedding module was $6514$. For the partitioning module, \cref{alg:partitioning_module}, we used $\nu_{\textnormal{pre}} = 2$, $\nu_{\textnormal{coarse}} = 1$, $\nu_{\textnormal{post}} =2 $, $\nu_{\textnormal{lin}} = 4$, all with $16$ input and output channels except for $\textnormal{SAGEConv}_{(\textnormal{first})}$ and $\text{Linear}_{\nu_{\text{lin}}}$, with the first having $1$ input and $16$ output channels and the last $16$ input and $2$ output channels, for a total of $3538$ tunable parameters. Training for both the embedding and partitioning module was performed using stochastic gradient descent with the Adam optimizer~\cite{adam2015} with learning rate $\delta = 10^{-3}$ and batch size $5$. All the above parameters were chosen after performing some tuning. We noticed that the training was not sensitive to the choice of the random seed.

As in~\cite{nazi2019gapICLR}, we refer to the final partitioning algorithm as described in \cref{sec:algorithm} -- the combination of the embedding and partitioning modules -- as the generalized approximate partitioning (GAP) framework. The GAP method, both the embedding and partitioning module, ran on the GPU using PyTorch and PyTorch Geometric, while all other codes ran on the CPU.

\subsection{Embedding Module}

\paragraph{Training}
We trained the embedding module on a variety of graphs, including $1500$ random Delaunay meshes on the square $[0,1]^2$, $750$ random Delaunay meshes on the rectangle $[0,2]\times[0,1]$, Finite Element (FEM) triangulations of different planar geometries called GradedL, Hole3 and Hole6 -- each with multiple levels of refinement, and several connected graphs from symmetric $2$D/$3$D discretizations from the SuiteSparse database \cite{suitesparse}. See \cref{app:additional_material}, \cref{fig:hole3_hole6_partitioning} and \cref{fig:gradedl_partitioning}, for illustrations of the GradedL, Hole3 and Hole6 finite element graphs. All graphs used to train the embedding module had between $100$ and $5000$ nodes. Both the FEM and the SuiteSparse graphs were oversampled -- $15$ times for FEM and $3$ times for SuiteSparse -- in order to enrich the training set -- resulting in $570$ FEM and $351$ SuiteSparse training graphs. The total number of graphs used for training was $3171$ and training was performed for $120$ epochs. Although some graphs were oversampled, training was different each time due to the non-deterministic coarsening. The training of this module took about 6 hours on our machine.

\paragraph{Evaluation}
In order to evaluate the quality of the trained embedding, we used the output from the embedding module (the approximate Fiedler vector) to partition a set of test graphs. We compared the quality of these partitions with spectral partitioning with the exact Fiedler vector, computed using ARPACK~\cite{lehoucq1998arpack} through Scipy. 
Partitioning with an (approximate) eigenvector was explained in \cref{ssec:spectralgraphpartitioning}. The set of graphs used for this evaluation was different than the training dataset; it included different discretization domains, with very different sparsity patterns, and much larger graphs. 
\Cref{tab:graphs_data} shows the number of graphs and the maximum number of nodes and edges in the graphs in each class. Since the coarsening was not deterministic, we evaluated the embedding module network twice and recorded the smallest normalized cut. The results are summarized in \crefrange{tab:delaunay_data}{tab:suitesparse_data},
which show the median of the normalized cut, the balance, the cut, and the runtime (in seconds) for each class of graphs, for the different methods. Looking at the lines \emph{App.~Spec} and \emph{Spectral},
one can see that the partitioning quality -- in terms of the normalized cut, using approximated spectral partitioning or exact spectral partitioning -- is comparable, and similar behavior is seen for the cut. Partition balances are close for all classes, with a slightly higher imbalance for the approximate spectral method for Hole3, Hole6 and SuiteSparse. Somewhat surprisingly, the class with the worst performance is Delaunay.  Approximated spectral partitioning produced larger normalized cut (by about 7\%) and cut (by about 6\%) than spectral partitioning, and it also resulted in a higher imbalance (by about 11\%).
Note also that the approximate spectral embedding was significantly faster than computing the true Fiedler vector. However, true spectral partitioning can be optimized further, for instance with a state-of-the-art multigrid preconditioned LOBPCG eigensolver running on the GPU~\cite{zhuzhunashvili2017preconditioned}.

\begin{table}
   \centering
   \begin{subtable}[t]{0.48\textwidth}
    \centering
    \small{
    \begin{tabular}{r|c|c|c}
         Class & \#graphs & max $|V|$ & max $|E|$ \\
         \hline
         Delaunay & $100$ & $300$K & $1.8$M \\
         Graded L & $98$ & $150$K & $1$M \\
         Hole3 & $99$ & $300$K & $1.7$M \\
         Hole6 & $99$ & $300$K & $1.7$M \\
         SuiteSparse & $162$ & $1$M & $10$M 
    \end{tabular}}
    \caption{Number of graphs, maximum number of nodes and edges for each class of graphs used for testing \label{tab:graphs_data}}
    \end{subtable}\vspace{0.3cm}
    \begin{subtable}[t]{0.48\textwidth}
    \centering
    \small{
    \begin{tabular}{r|c|c|c|c}
         \textbf{Delaunay} & NC & B & C & T \\
         \hline
         GAP & $0.003$ & $1.15$ & $534.5$ & $0.5994$ \\
         METIS & $0.0026$ & $1$ & $441.5$ & $3.5897$ \\
         Scotch & $0.0026$ & $1.01$ & $449$ & $0.098$ \\
         App. Spec & $0.003$ & $1.25$ & $487.5$ & $0.2644$ \\
         Spectral & $0.0028$ & $1.11$ & $453$ & $26.5136$ 
    \end{tabular}}
    \caption{Delaunay triangulations\label{tab:delaunay_data}}
    \end{subtable}\vspace{.3cm}
    \begin{subtable}[t]{0.48\textwidth}
    \centering
    \small{
    \begin{tabular}{r|c|c|c|c}
         \textbf{Graded L} & NC & B & C & T \\
         \hline
         GAP & $0.007$ & $1.15$ & $410$ & $0.2336$ \\
         METIS & $0.007$ & $1$ & $425$ & $0.6223$ \\
         Scotch & $0.007$ & $1.01$ & $429$ & $0.0143$ \\
         App. Spec & $0.0065$ & $1.82$ & $369$ & $0.101$ \\
         Spectral & $0.0063$ & $1.9$ & $350$ & $1.7228$
    \end{tabular}}
    \caption{Graded L triangulations\label{tab:gradedl_data}}
    \end{subtable}\vspace{0.3cm}
    \begin{subtable}[t]{0.48\textwidth}
    \centering
    \small{
    \begin{tabular}{r|c|c|c|c}
         \textbf{Hole3} & NC & B & C & T \\
         \hline
         GAP & $0.002$ & $1.20$ & $223.5$ & $0.3142$ \\
         METIS & $0.002$ & $1$ & $219.5$ & $1.1158$ \\
         Scotch & $0.0023$ & $1.01$ & $249.5$ & $0.0186$ \\
         App. Spec & $0.0019$ & $1.14$ & $208$ & $0.1348$ \\
         Spectral & $0.002$ & $1.08$ & $212$ & $7.9833$  
    \end{tabular}}
    \caption{Hole3 triangulations\label{tab:hole3_data}}
    \end{subtable} \\ \vspace{0.3cm}
    \begin{subtable}[t]{0.48\textwidth}
    \centering
    \small{
    \begin{tabular}{r|c|c|c|c}
         \textbf{Hole6} & NC & B & C & T \\
         \hline
         GAP & $0.0027$ & $1.13$ & $341$ & $0.3135$ \\
         METIS & $0.0024$ & $1$ & $314$ & $1.1344$ \\
         Scotch & $0.0026$ & $1.01$ & $345$ & $0.02$ \\
         App. Spec & $0.0025$ & $1.08$ & $315$ & $0.1358$ \\
         Spectral & $0.0023$ & $1$ & $311.5$ & $7.3896$ 
    \end{tabular}}
    \caption{Hole6 triangulations\label{tab:hole6_data}}
    \end{subtable}\vspace{0.3cm}
    \begin{subtable}[t]{0.48\textwidth}
    \centering
    \small{
    \begin{tabular}{r|c|c|c|c}
         \textbf{SuiteSparse} & NC & B & C & T \\
         \hline
         GAP & $0.0125$ & $1.14$ & $2859$ & $0.3944$ \\
         METIS & $0.0128$ & $1.01$ & $3270$ & $0.7179$ \\
         Scotch & $0.0127$ & $1.01$ & $3096$ & $0.0766$ \\
         App. Spec & $0.011$ & $1.26$ & $2106$ & $0.172$ \\
         Spectral & $0.0098$ & $1.18$ & $1941$ & $1.2393$ 
    \end{tabular}}
    \caption{SuiteSparse graphs\label{tab:suitesparse_data}}
    \end{subtable}
    \caption{The median of the normalized cut (NC), balance (B), cut (C) and runtime (T) in seconds for $5$ different datasets. \Cref{tab:graphs_data} describes the datasets.\label{tab:all_data}}
\end{table}
\begin{table*}
   \centering
   \small{
        \begin{tabular}{c|c|c|c|c|c|c|c|c} 
       Graph & Group & Nodes & Edges & $\text{NC}_{\text{G}}$ & $\text{NC}_{\text{M}}$ & $\text{NC}_{\text{S}}$ & $\text{NC}_{\text{ASp}}$ & $\text{NC}_{\text{Sp}}$  \\ \hline
       ML\_Geer & Janna & $1504002$ & $110686677$ & $0.0019$ & $0.0021$ & $0.002$ & $0.0019$ & $0.0019$ \\
       xenon2 & Ronis & $157410$ & $3709224$ & $0.0127$ & $0.0122$ & $0.0125$ & $0.013$ & $0.0116$ \\
       torso3 & Norris & $259156$ & $4429042$ & $0.013$ & $0.0132$ & $0.0123$ & $0.0119$ & $0.0073$ \\
        Transport & Janna & $1601150$ & $21885170$ & $0.0036$ & $0.0036$ & $0.0036$ & $0.0036$ & $0.0036$ \\ 
       stomach & Norris & $213360$ & $3021648$ & $0.0029$ & $0.0031$ & $0.0027$ & $0.0027$ & $0.0026$ \\
        Flan\_1565 & Janna & $1564794$ & $114165372$ & $0.0012$ & $0.0013$ & $0.0012$ & $0.0012$ & $0.0012$\\
        dielFilterV3real & Dziekonski & $1102824$ & $89306020$ & $0.003$ & $0.0028$ & $0.0028$ & $0.003$ & $0.0028$\\
        af\_shell10 & Schenk\_AFE & $1508065$ & $52259885$ & $0.0021$ & $0.0022$ & $0.0025$ & $0.0021$ & $0.0021$\\
        CurlCurl\_3 & Bodendiek & $1219574$ & $13544618$ & $0.0086$ & $0.0081$ & $0.0083$ & $0.0035$ & $0.0027$
    \end{tabular}}
    \caption{Comparison of the normalized cut obtained with GAP ($\text{NC}_{\text{G}}$), METIS ($\text{NC}_{\text{M}}$), Scotch ($\text{NC}_{\text{S}}$), Approximate spectral partitioning ($\text{NC}_{\text{ASp}}$) and Spectral partitioning ($\text{NC}_{\text{Sp}}$) for several graphs taken from the SuiteSparse database. The columns ``Nodes'' and ``Edges'' display the number of nodes and edges of the largest connected component of the graph.}\label{tab:janna}
\end{table*}

\subsection{Partitioning Module}
\paragraph{Training}
The training dataset for the partitioning module was similar to that used for training the embedding module, but with smaller graphs. More precisely, the dataset included $20$ random Delaunay meshes on the square $[0,1]^2$, $10$ random Delaunay meshes on the rectangle $[0,2]\times[0,1]$, the FEM triangulations described above, and $27$ connected graphs from symmetric 2D/3D discretizations from the SuiteSparse database, all with number of nodes between $100$ and $500$. The FEM set was oversampled $5$ times so there was a total of $20$ triangulations. The total number of graphs used for training was $77$ and we trained for $500$ epochs. The training of this module took less than 1 hour on our machine.

We tested the trained model on the same Delaunay, FEM triangulations and SuiteSparse matrices used to test the embedding module. Again, since the coarsening was not deterministic, we evaluated the network twice and kept the partitioning with the smallest normalized cut. \Cref{tab:delaunay_data} to \cref{tab:suitesparse_data} compare the median normalized cut, balance, cut and runtime to that from METIS, Scotch, approximate spectral and exact spectral graph partitioning for each class of graphs. We also considered some bigger possibly non-connected and non-symmetric SuiteSparse discretizations, whose normalized cuts obtained with the $5$ different methods were collected in \cref{tab:janna}. For the non-symmetric ones we first made the adjacency matrix $A$ symmetric by considering $\frac{1}{2}(A+A^T)$.

Both METIS and Scotch minimize a different objective than GAP. More precisely, they aim to minimize the cut while keeping the partitions of approximately the same \emph{cardinality}. The default tolerance for the imbalance for METIS is $1.01$. However, for our GAP approach we still decide to minimize the normalized cut, which already takes into account the imbalance of the partition \emph{volumes}, regardless of the cardinality imbalance. As discussed in \cref{ssec:partitioning_module}, the GAP loss function \cref{eq:gap_loss} can be modified to include a term for the (cardinality) imbalance, see \cref{eq:gap_loss_balanced}.

\paragraph{Evaluation}
We see that for the FEM and SuiteSparse graphs the normalized cut and the cut are very close for the different methods, with GAP outperforming METIS and Scotch on the SuiteSparse dataset. The GAP partitions have higher imbalance than METIS and Scotch, but overall the peak imbalance is $20\%$ for the Hole3 class, while for Graded L we observe that GAP has much lower imbalance than approximate spectral and spectral partitioning, meaning that the model tried to correct the high imbalance of the spectral methods. For a pictorial representation of this phenomenon, see \cref{fig:gradedl_partitioning} in \cref{app:additional_material}. The Delaunay class is the one in which GAP has normalized cut and cut higher than the other methods. Regarding the runtimes, note that GAP consistently outperforms both METIS and spectral partitioning, while being slower than approximate spectral, which only requires a single neural network (the embedding module), and Scotch. Scotch is fastest in $93\%$ of cases. METIS was called through the NetworkX-METIS interface\footnote{https://networkx-metis.readthedocs.io/en/latest/}, while for Scotch we wrote a Python ctypes interface\footnote{Both METIS and Scotch were run on the CPU. Through our Python interface we only had access to the sequential versions of METIS and Scotch.}. We can also observe that the GAP model generalizes very well to different and much bigger graphs than the ones seen in the training, like the ones in \cref{tab:janna}, where it often outperforms METIS and Scotch. Recall that the test graphs often exhibited very different sparsity patterns than the ones seen during training.

\begin{figure}
    \centering
    \begin{subfigure}[t]{.48\textwidth}
      \centering
      \includegraphics[width=\textwidth]{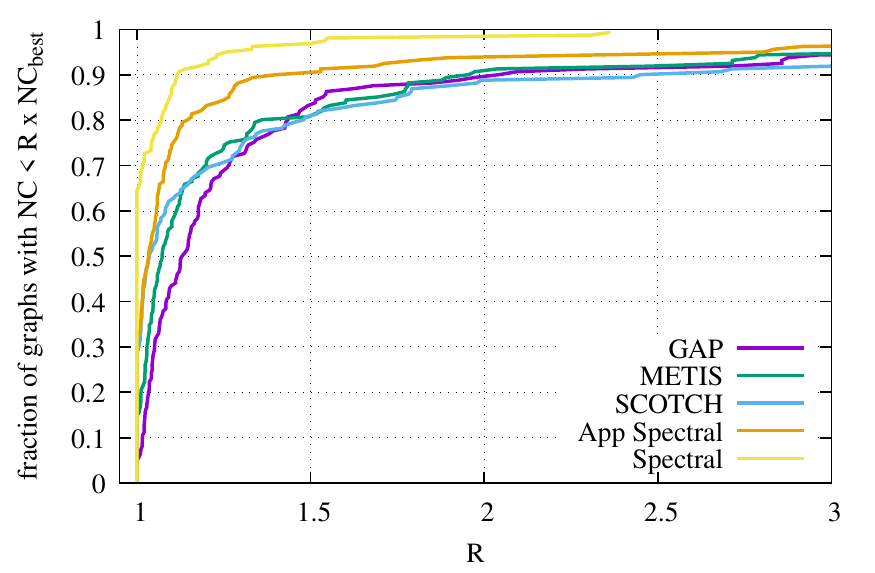}
      \caption{Normalized cut\label{fig:pp_nc_ss}}
    \end{subfigure}
    \hfill
    \begin{subfigure}[t]{.48\textwidth}
      \centering
      \includegraphics[width=\textwidth]{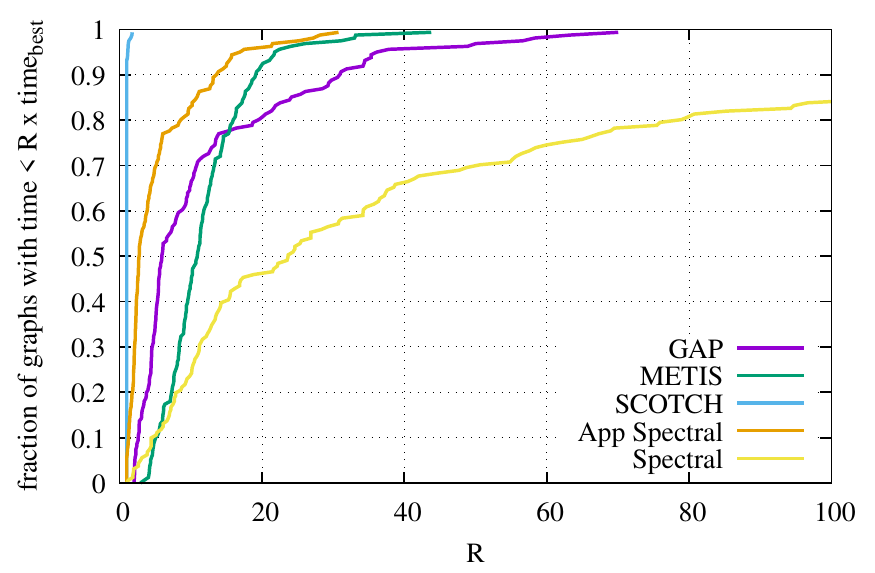}
      \caption{Time\label{fig:pp_time_ss}}
    \end{subfigure}
    \caption{Performance profiles comparing the normalized cut (\Cref{fig:pp_nc_ss}) and the runtime (\Cref{fig:pp_time_ss}) for the different partitioners on the SuiteSparse dataset. Vertical axis shows the fraction of the graphs for which a method is within a ratio $R$ (horizontal axis) of the best method for that graph.}
    \label{fig:ss_performace}
\end{figure}

\begin{figure}
    \centering
    \begin{subfigure}[t]{0.425\textwidth}
    \centering
      \includegraphics[width=\textwidth]{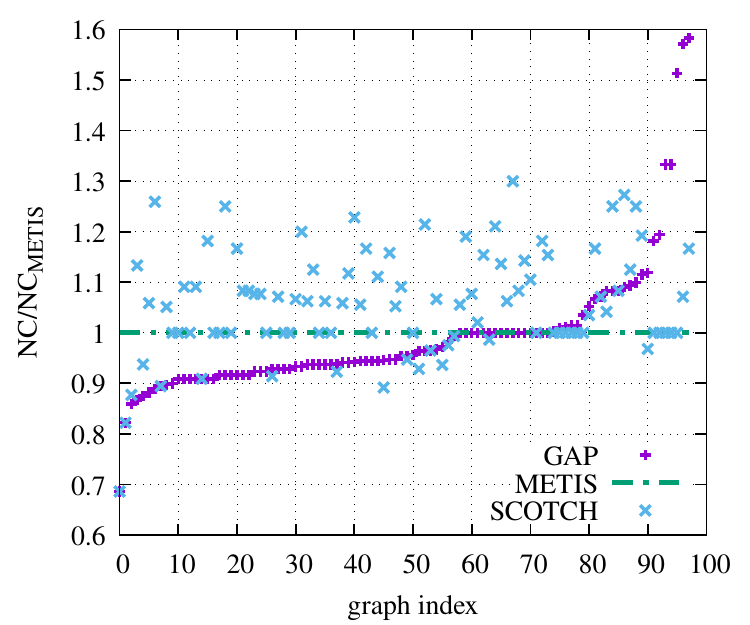}
      \caption{Hole3 normalized cut\label{fig:scatter_Hole3}}
    \end{subfigure}
    \hfill
    \begin{subfigure}[t]{0.425\textwidth}
    \centering
      \includegraphics[width=\textwidth]{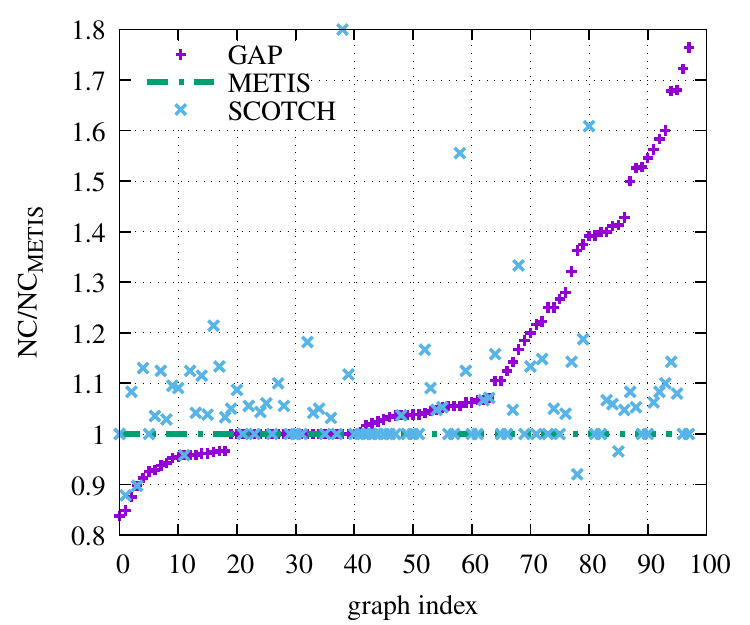}
      \caption{Hole6 normalized cut\label{fig:scatter_Hole6}}
    \end{subfigure}
    \hfill
    \begin{subfigure}[t]{0.425\textwidth}
      \centering
      \includegraphics[width=\textwidth]{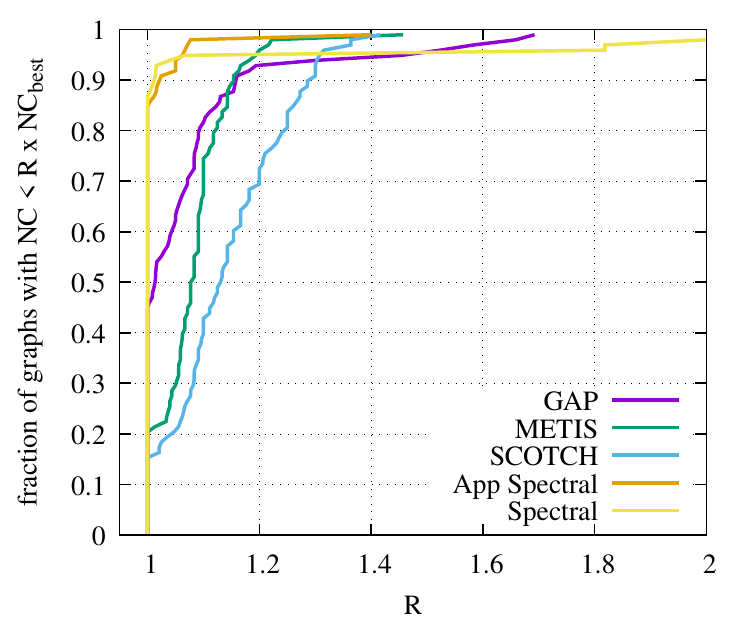}
      \caption{Hole3 performance profile\label{fig:scatter_Hole3_pp}}
    \end{subfigure}
    \caption{Comparison of the normalized cut for the Hole3 and Hole6 datasets. \Crefrange{fig:scatter_Hole3}{fig:scatter_Hole6} show the normalized cut relative to the normalized cut obtained with METIS. The graphs are sorted based on the normalized cut obtained with GAP (relative to METIS). \Cref{fig:scatter_Hole3_pp} is a performance profile comparing the different methods on the Hole3 dataset.}
    \label{fig:scatter_plots_hole3_hole6}
\end{figure}

\Cref{fig:ss_performace} shows performance profiles~\cite{dolan2002benchmarking} for the normalized cut and the runtime for the SuiteSparse datasets, while \cref{fig:scatter_plots_hole3_hole6} compares the normalized cuts for the Hole3 and Hole6 datasets. In \cref{fig:scatter_Hole3,fig:scatter_Hole6}, the normalized cut was normalized with respect to the normalized cut obtained with METIS, and sorted based on the GAP results. \Cref{fig:scatter_Hole3_pp} shows a performance profile comparing the different methods on the Hole3 dataset. This shows the fraction of problems for which a given method is within a ratio $R$ (horizontal axis) of the best method. Hence, higher and more to the left means better.

\section{Conclusions and Outlook\label{sec:conclusions}}
We presented a graph bisection algorithm, for graphs without node or edge features, based on graph neural networks. The neural network consisted of two parts, an embedding and a partitioning module. The embedding module was trained to output an approximation to the Fiedler vector, and thus could be used to perform approximate spectral partitioning. The quality of this approximate spectral partitioning was very close to the exact spectral partitioning, while being considerably faster. This approximation algorithm also turned out to generalize very well to graphs that were much larger, and with different sparsity patterns, than the graphs in the training dataset. Note that the SuiteSparse collection contains a wide variety of problems. We focused on graphs coming from 2D/3D discretizations, since graphs coming from other applications may exhibit very different partitioning properties.

Note that the spectral embedding was passed to a second neural network, which was trained to minimize the normalized cut. Currently we only pass a single approximate eigenvector to the partitioning module. Likely, the partitioning quality could be improved by passing a higher order spectral embedding to the partitioning module~\cite{alpert1999spectral}. Alternatively, we will also experiment with enhancing the feature tensor with other features, such as even random vectors, which have been shown to be able to improve performance of graph neural networks~\cite{sato2021random}. In any case, for K-way partitioning ($g > 2$), more features will be required, or which could be trained to minimize related objectives, e.g.,~\cref{eq:gap_loss_balanced}.

We are confident that the structure of both neural networks can be tuned further. We believe it should be possible to formulate a leaner network for the partitioning module, for instance without the multilevel aspect since the embedding already provides a global view of the graph. This would improve the performance of the overall partitioning algorithm.

\section*{Acknowledgements}
This work was supported by the Laboratory Directed Research and Development Program of Lawrence Berkeley National Laboratory under U.S. Department of Energy Contract No. DE-AC02-05CH11231.

\newpage
\bibliographystyle{plain}
\bibliography{refs}

\newpage

\appendix
\section{Supplementary Material}\label{app:additional_material}

\subsection{Loss function implementation}

\Cref{fig:gap_loss} shows an efficient implementation of the loss function presented in \cite{nazi2019gapICLR} using the Pytorch and Pytorch Geometric tools. \texttt{graph} is a \texttt{torch\_geometric.Data} object, with the connectivity stored in the COO format in \texttt{graph.edge\_index}, a $2 \times |E|$ tensor.

\begin{figure}[h]
\begin{lstlisting}[language=Python]
import torch
from torch_geometric.utils import degree
def loss_normalized_cut(y, graph):
    d = degree(graph.edge_index[0],                num_nodes=y.size(0))
    gamma = torch.t(y) @ d
    c = torch.sum(y[graph.edge_index[0], 0] * 
        y[graph.edge_index[1], 1])
    return torch.sum(torch.div(c, gamma))
\end{lstlisting}
  \caption{Implementation of the loss function using PyTorch and Pytorch Geometric. \label{fig:gap_loss}}
\end{figure}

\subsection{Finite Element domains with holes}
\begin{figure}[h]
    \centering
    \begin{subfigure}[t]{0.48\textwidth}
    \centering
      \includegraphics[scale=0.4]{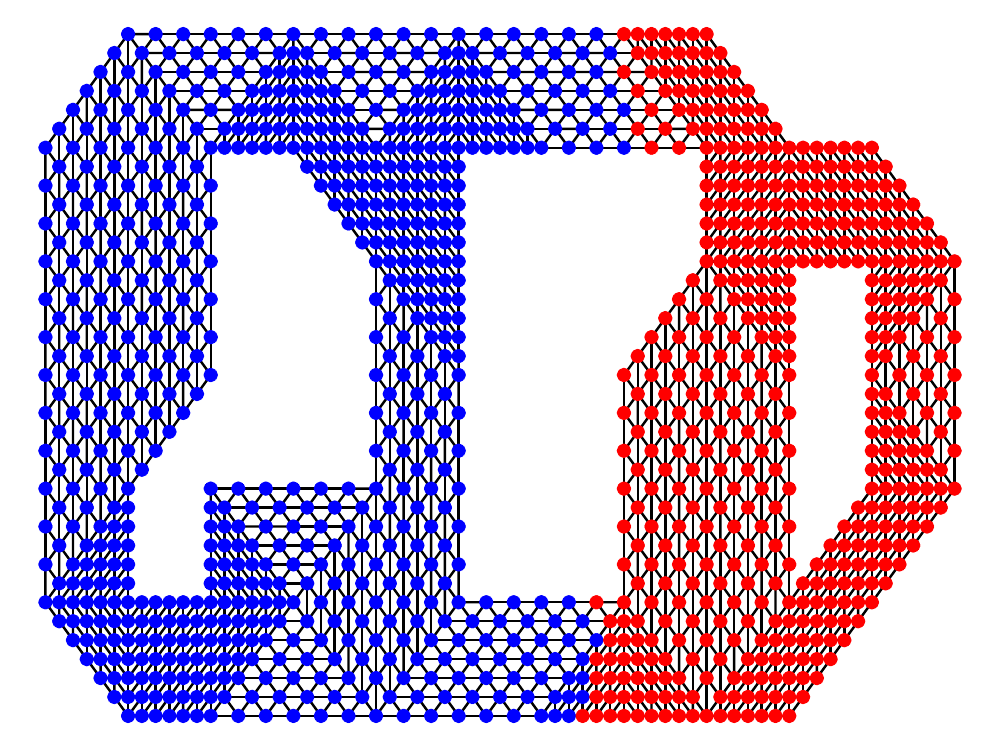}
    \end{subfigure}
     \hfill
    \begin{subfigure}[t]{0.48\textwidth}
    \centering
      \includegraphics[scale=0.4]{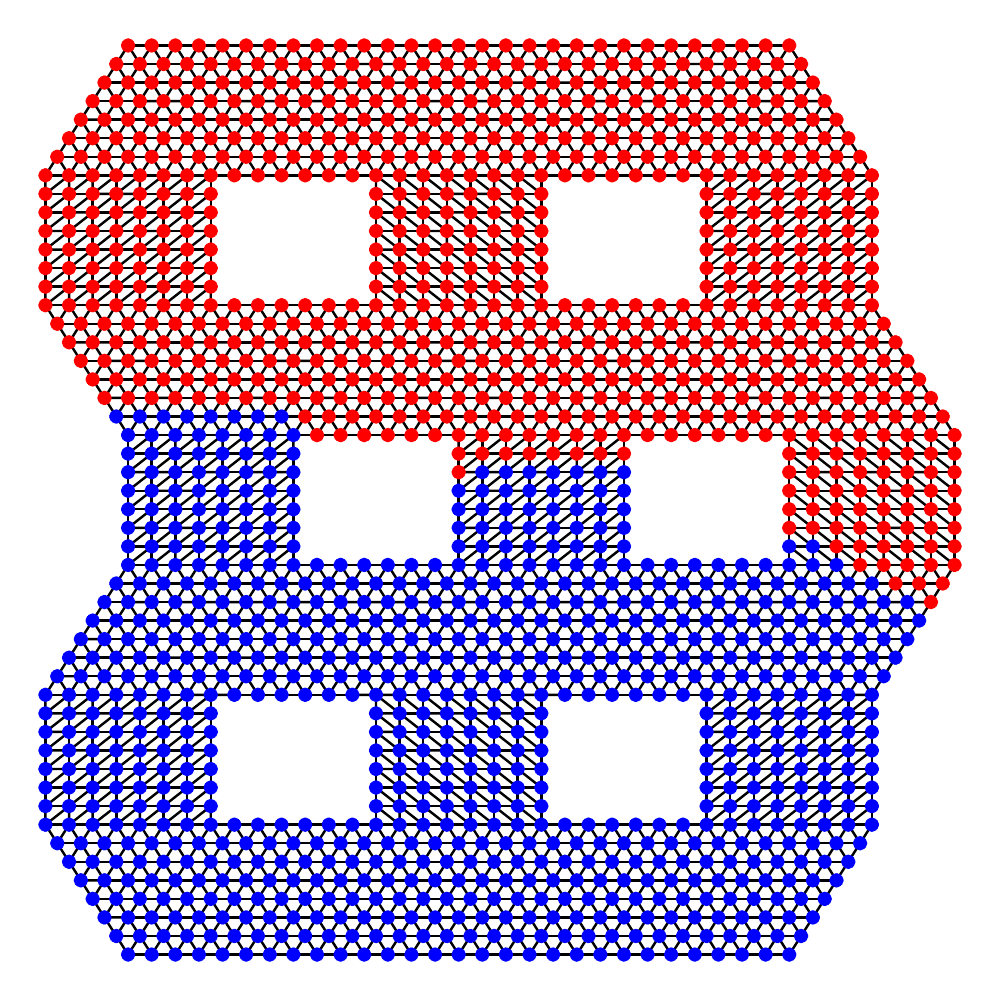}
    \end{subfigure}
    \caption{Illustration of the GAP partitioning for two Finite Element triangulations. On the top, Hole3 after $6$ refinements and on the bottom Hole6 after $7$ refinements.}
    \label{fig:hole3_hole6_partitioning}
\end{figure}

\Cref{fig:hole3_hole6_partitioning} illustrates the GAP partitioning for two Finite Element triangulations containing holes. More precisely, the Hole3 domain after $6$ refinements and Hole6 after 7 refinements. For both meshes we note that the GAP model takes the advantage of the existing holes to reduce the edge cut.

\subsection{Finite Element L-shaped domain}
\begin{figure}[h]
    \centering
    \includegraphics[scale=0.6]{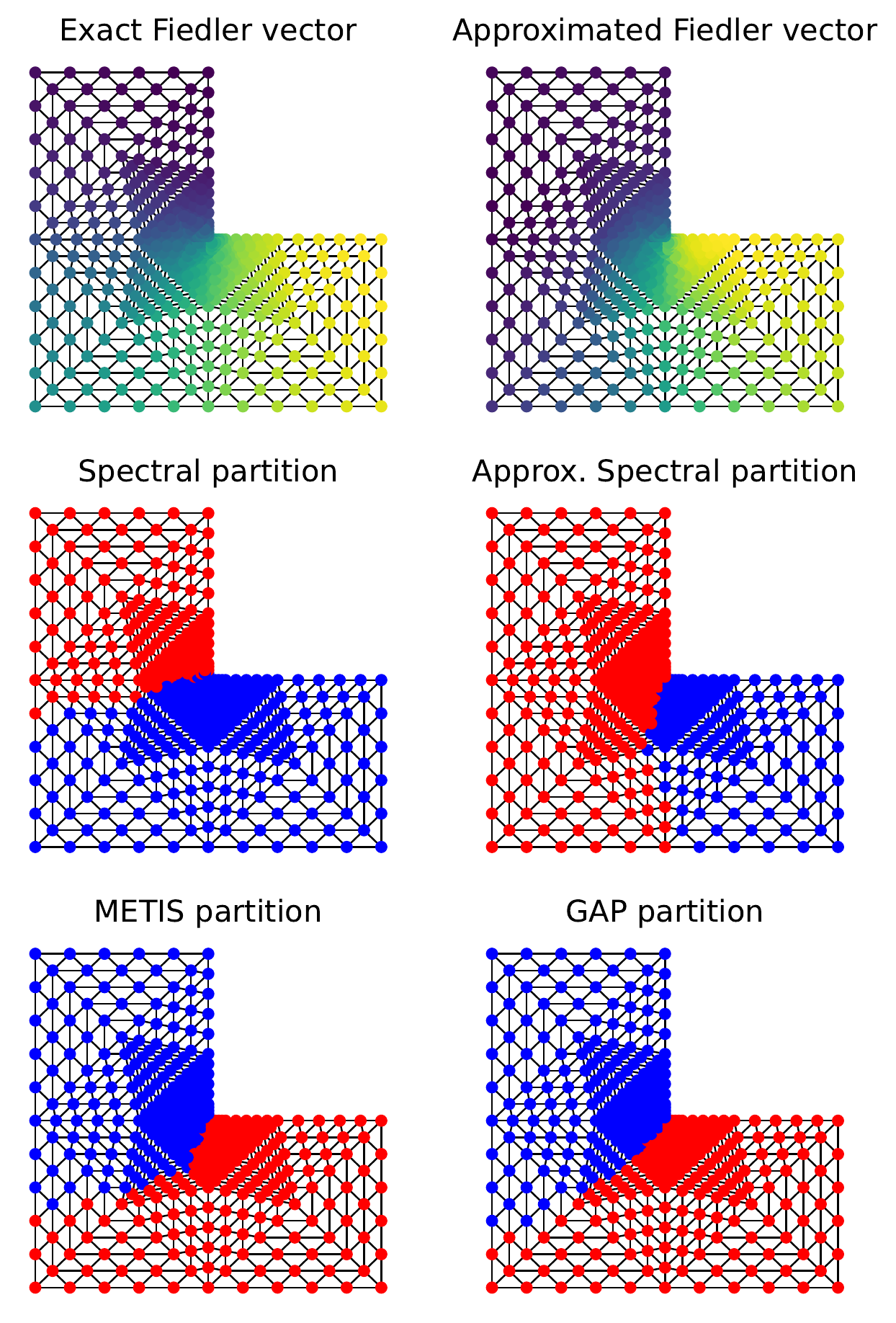}
    \caption{Fiedler vector and partitioning on the Graded L Finite Element triangulation after $5$ refinements. Top row: exact Fiedler vector (left), approximated one (right). Middle row: spectral partitionings associated with the exact Fiedler vector (left) and the approximated one (right). Bottom row: METIS partitioning (left) and GAP partitioning (right).}
    \label{fig:gradedl_partitioning}
\end{figure}

\Cref{fig:gradedl_partitioning} illustrates the Fiedler vector and the approximated one, the associated spectral partitionings and the METIS and GAP partitions. Note that both spectral methods cut vertically or horizontally the L-shaped domain, providing small cut but also unbalanced partitions. On the other hand, both GAP and METIS cut close the diagonal between the inner and the outer corner of the L. This also produces small cut, but also more balanced partitions.

\end{document}

%% file: partition_fig.tex
\begin{figure*}
\centering
\begin{subfigure}[t]{1\textwidth}
  \centering
  \vskip 0pt
  \resizebox{\textwidth}{!}{
  \begin{tikzpicture}[yscale=0.8,xscale=1.3]
    \node[anchor=west] at (0,0) (nodeG0) {$G^0$};
    \node[anchor=west] at (.5,-1.5) (nodeG1) {$G^1$};
    \node[anchor=west] at (1,-3) (nodeG2) {$G^2$};
    \node[anchor=west] at (1.2,-4.5) (nodeG3) {$G^3, F^3$};
    \node at (1.5,-5.3) (nodeG3_F0) {$F^3 = \begin{bmatrix} 1 & 0 \\ 0 & 1 \end{bmatrix}$};

    \draw[thick, ->] (nodeG0) -- node [midway] {pool} (nodeG1);
    \draw[thick, ->] (nodeG1) -- node [midway] {pool} (nodeG2);
    \draw[thick, ->] (nodeG2) -- node [midway] {pool} (nodeG3);

    \node[anchor=west] at (3.5,-4.5) (nodeG3_2) {$G^3, F^3$};
    \node[anchor=west] at (4,-3) (nodeG2_2) {$G^2, F^2$};
    \node[anchor=west] at (6,-3) (nodeG2_3) {$G^2, F^2$};
    \node[anchor=west] at (6.5,-1.5) (nodeG1_2) {$G^1, F^1$};
    \node[anchor=west] at (8.5,-1.5) (nodeG1_3) {$G^1, F^1$};
    \node[anchor=west] at (9,0) (nodeG0_2) {$G^0, F^0$};
    \node[anchor=west] at (11,0) (nodeG0_3) {$G^0, F^0$};
    \node[anchor=west] at (13,0) (nodeG0_4) {$G^0, F^0$};

    \draw[thick, ->] (nodeG3) -- node [midway, above] {conv} (nodeG3_2);
    \draw[thick, ->] (nodeG3_2) -- node [midway] {unpool} (nodeG2_2);
    \draw[thick, ->] (nodeG2_2) -- node [midway, above] {conv} (nodeG2_3);
    \draw[thick, ->] (nodeG2_3) -- node [midway] {unpool} (nodeG1_2);
    \draw[thick, ->] (nodeG1_2) -- node [midway, above] {conv} (nodeG1_3);
    \draw[thick, ->] (nodeG1_3) -- node [midway] {unpool} (nodeG0_2);
    \draw[thick, ->] (nodeG0_2) -- node [midway, above] {conv} (nodeG0_3);
    \draw[thick, ->] (nodeG0_3) -- node [midway, above] {dense+QR} (nodeG0_4);
  \end{tikzpicture}
  }
  \caption{\centering Schematic view of the neural network for the embedding phase with $3$ levels of coarsening.}
  \label{fig:embedding_net}
  \vspace{0.1cm}
\end{subfigure}
\hfill
\begin{subfigure}[t]{1\textwidth}
  \centering
  \vskip 0pt
  \resizebox{\textwidth}{!}{
  \begin{tikzpicture}[yscale=0.8,xscale=1.3]  
    \node[anchor=west] at (0,0) (nodeG0) {$G^0$};
    \node[anchor=west] at (1,0) (nodeG0_0) {$G^0$};
    \node[anchor=west] at (2,-1.5) (nodeG1) {$G^1$};
    \node[anchor=west] at (3,-1.5) (nodeG1_0) {$G^1$};
    \node[anchor=west] at (4,-3) (nodeG2) {$G^2$};
    \node[anchor=west] at (5,-3) (nodeG2_0) {$G^2$};
    \node[anchor=west] at (6,-4.5) (nodeG3) {$G^3$};

    \draw[thick, ->] (nodeG0) -- node [midway, above] {conv} (nodeG0_0);
    \draw[thick, ->] (nodeG0_0) -- node [midway] {pool} (nodeG1);
    \draw[thick, ->] (nodeG1) -- node [midway, above] {conv} (nodeG1_0);
    \draw[thick, ->] (nodeG1_0) -- node [midway] {pool} (nodeG2);
    \draw[thick, ->] (nodeG2) -- node [midway, above] {conv} (nodeG2_0);
    \draw[thick, ->] (nodeG2_0) -- node [midway] {pool} (nodeG3);

    \node[anchor=west] at (7,-4.5) (nodeG3_2) {$G^3$};
    \node[anchor=west] at (8,-3) (nodeG2_2) {$G^2$};
    \node[anchor=west] at (9,-3) (nodeG2_3) {$G^2$};
    \node[anchor=west] at (10,-1.5) (nodeG1_2) {$G^1$};
    \node[anchor=west] at (11,-1.5) (nodeG1_3) {$G^1$};
    \node[anchor=west] at (12,0) (nodeG0_2) {$G^0$};
    \node[anchor=west] at (13,0) (nodeG0_3) {$G^0$};
    \node[anchor=west] at (14,0) (nodeG0_4) {$G^0$};

    \draw[thick, ->] (nodeG3) -- node [midway, above] {conv} (nodeG3_2);
    \draw[thick, ->] (nodeG3_2) -- node [midway] {unpool} (nodeG2_2);
    \draw[thick, ->] (nodeG2_2) -- node [midway, above] {conv} (nodeG2_3);
    \draw[thick, ->] (nodeG2_3) -- node [midway] {unpool} (nodeG1_2);
    \draw[thick, ->] (nodeG1_2) -- node [midway, above] {conv} (nodeG1_3);
    \draw[thick, ->] (nodeG1_3) -- node [midway] {unpool} (nodeG0_2);
    \draw[thick, ->] (nodeG0_2) -- node [midway, above] {conv} (nodeG0_3);
    \draw[thick, ->] (nodeG0_3) -- node [midway, above] {dense} (nodeG0_4);

    \draw[dashed, ->] (nodeG0_0) -- node [midway, above] {residual connection} (nodeG0_2);
    \draw[dashed, ->] (nodeG1_0) -- (nodeG1_2);
    \draw[dashed, ->] (nodeG2_0) -- (nodeG2_2);
  \end{tikzpicture}
  }
  \caption{\centering Schematic view of the neural network for the partitioning phase with $3$ levels of coarsening.}
  \label{fig:partition_net}
\end{subfigure}
  \caption{Illustration of the neural network used for partitioning. The embedding module from \Cref{fig:embedding_net}, see \Cref{ssec:spectral_feature}, is trained first using \Cref{eq:eig_loss} as the loss. Next, the partitioning module in \Cref{fig:partition_net}, see \Cref{ssec:partitioning_module}, is trained using the \Cref{eq:gap_loss} as the loss.}
  \label{fig:embedding_partition_net}
\end{figure*}